\documentclass[10pt,twocolumn,letterpaper]{article}

\usepackage[pagenumbers]{cvpr}

\newcommand{\tablestyle}[2]{\setlength{\tabcolsep}{#1}\renewcommand{\arraystretch}{#2}\centering}

\usepackage{multirow}
\usepackage{multicol}
\usepackage{amssymb}% http://ctan.org/pkg/amssymb
\usepackage{pifont}% http://ctan.org/pkg/pifont
\usepackage{adjustbox} 
\usepackage{subcaption}

\definecolor{citeblue}{rgb}{0.21,0.49,0.74}
\usepackage[pagebackref,breaklinks,colorlinks,allcolors=citeblue]{hyperref}

\newcommand{\ours}{\mbox{PixelFlow}\xspace}

\title{PixelFlow: Pixel-Space Generative Models with Flow}

\author{Shoufa Chen$^1$ \quad Chongjian Ge$^{1,2}$ \quad Shilong Zhang$^1$ \quad Peize Sun$^1$ \quad Ping Luo$^1$
\vspace{2pt} \\
$^1$The University of Hong Kong \quad $^2$Adobe
}

\begin{document}

\maketitle

\begin{abstract}
We present PixelFlow, a family of image generation models that operate directly in the raw pixel space, in contrast to the predominant latent-space models. This approach simplifies the image generation process by eliminating the need for a pre-trained Variational Autoencoder (VAE) and enabling the whole model end-to-end trainable. Through efficient cascade flow modeling, PixelFlow achieves affordable computation cost in pixel space. It achieves an FID of \textbf{1.98} on 256$\times$256 ImageNet class-conditional image generation benchmark. The qualitative text-to-image results demonstrate that \ours excels in image quality, artistry, and semantic control. We hope this new paradigm will inspire and open up new opportunities for next-generation visual generation models. Code and models are available at 
\small{\url{https://github.com/ShoufaChen/PixelFlow}}.
\end{abstract}
    
\section{Introduction}\label{sec:intro}
\vspace{3mm}
\textit{\quad \quad Numquam ponenda est pluralitas sine necessitate.}
\begin{flushright}
— \textit{\normalsize{William of Ockham }}
\vspace{3mm}
\end{flushright}

\begin{figure}[t]
\centering
\includegraphics[width=0.96\linewidth]{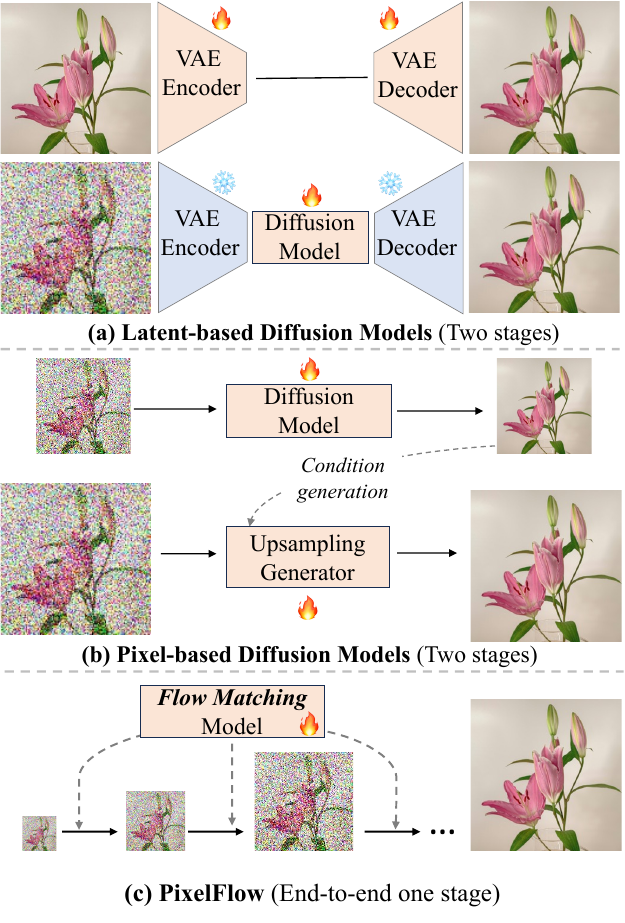}
\caption{\textbf{Comparisons of Design Paradigms} between latent-based diffusion models~(LDMs), pixel-based diffusion models~(PDMs), and \ours: (a) LDMs split training into two separate stages—first independently training off-the-shell VAEs, then training diffusion models on tokens extracted from the pre-trained VAEs; (b) Previous PDMs typically train two separate models: a diffusion model on low-resolution images and an upsampler for high-resolution synthesis; (c) \ours, by contrast, offers an end-to-end solution for pixel-based generation, combining both high efficiency and strong generative performance.}\label{fig:teaser_comparisons}
\vspace{-2pt}
\end{figure}

Driven by the success of the Stable Diffusion (SD) model series~\citep{rombach2022high,podell2023sdxl,stablecascade, esser2024scaling}, latent diffusion models~(LDMs)~\cite{rombach2022high} have emerged as the \textit{de facto} standard for generative modeling across diverse modalities, spanning image~\cite{esser2024scaling,peebles2023scalable,flux2024}, video~\citep{hong2022cogvideo,yang2024cogvideox,chen2024gentron,chen2025goku,zhang2025flashvideo}, audio~\citep{liu2023audioldm,evans2024fast}, and 3D~\citep{stan2023ldm3d, zeng2022lion}. As shown in \Cref{fig:teaser_comparisons}~(a), LDMs compress raw data into a compact latent space using pre-trained Variational Autoencoders (VAEs). This compression reduces computational demands and facilitates efficient diffusion denoising. Despite their widespread success, LDMs decouple the VAE and diffusion components, hindering joint optimization and complicating holistic diagnosis.

An alternative approach is to implement diffusion models in the raw pixel space. While intuitive, this becomes computationally unaffordable for high-resolution images due to the substantial resources required to process per-pixel correlations. Considering this, prior research~\citep{saharia2022image, gu2023matryoshka, ho2022cascaded, saharia2022photorealistic, edifyimage} has typically adopted a cascaded approach: first generating a low-resolution image, then employing additional upsamplers to produce high-quality outputs, with the low-resolution image serving as conditioning input, as shown in \Cref{fig:teaser_comparisons}(b).
However, these cascaded methods also introduce separate networks for different stages, still limiting the benefits of end-to-end design.

In this work, we introduce \ours, a simple but effective end-to-end framework for direct image generation in raw pixel space, without the need of separate networks like VAEs or upsamplers. As illustrated in \Cref{fig:teaser_comparisons}(c), \ours uses a unified set of parameters to model multi-scale samples across cascading resolutions via Flow Matching~\cite{liu2023flow, lipman2023flow}. At early denoising stages, when noise levels are high, \ours operates on lower-resolution samples. As denoising progresses, the resolution gradually increases until it reaches the target resolution in the final stage. This progressive strategy avoids performing all denoising steps at full resolution, thereby significantly reducing the overall computational cost of the generation process.

During training, the cross-scale samples at different timesteps are constructed by: 
(1) resizing the images to successive scales and adding Gaussian noise to each scaled image; (2) interpolating between adjacent scale noisy images as model input and conducting velocity prediction. The entire model is trained end-to-end using uniformly sampled training examples from all stages. During inference, the process begins with pure Gaussian noise at the lowest resolution. The model then progressively denoises and upscales the image until the target resolution is reached.

We evaluated \ours on both class-conditional and text-to-image generation tasks. Compared to established latent-space diffusion models~\cite{rombach2022high,peebles2023scalable,ma2024sit}, \ours delivers competitive performance. For instance, on the $256\times256$ ImageNet class-conditional generation benchmark, \ours achieves an FID of \textbf{1.98}. For text-to-image generation, \ours is evaluated on widely-used benchmarks, achieving \textbf{0.64} on GenEval~\cite{ghosh2024geneval} and \textbf{77.93} on DPG-Bench~\cite{hu2024ella_dbgbench}. In addition, qualitative results in \Cref{fig:t2i_512} and \Cref{fig:t2i_1024} illustrate that \ours has strong visual fidelity and text-image alignment, highlighting the potential of pixel-space generation for future research.

The \textbf{contributions} of \ours are summarized as in the following three points: 
\begin{itemize}
    \item By eliminating the need for a pre-trained VAE, we establish an end-to-end trainable image generation model in raw pixel space directly. 
    \item Through cascade flow modeling from low resolution to high resolution, our model achieves affordable computation cost in both training and inference.
    \item \ours obtains competitive performance in visual quality, including 1.98 FID on $256\times256$ ImageNet class-conditional image generation benchmark and appealing properties on text-to-image generation.
\end{itemize}

\section{Related Work}

\paragraph{Latent Space Diffusion/Flow Models.} 
Variational Autoencoders (VAEs) have become a core component in many recent generative models~\cite{esser2021taming,ramesh2021zero,rombach2022high,podell2023sdxl,sun2024autoregressive,esser2024scaling,yang2024cogvideox,flux2024}, enabling the mapping of visual data from pixel space to a lower-dimensional, perceptually equivalent latent space. This compact representation facilitates more efficient training and inference. However, VAEs often compromise high-frequency details~\cite{podell2023sdxl}, leading to inevitable low-level artifacts in generated outputs. Motivated by a desire for algorithmic simplicity and fully end-to-end optimization, we forgo the VAE and operate directly in pixel space.

\paragraph{Pixel Space Diffusion/Flow Models.} Early diffusion models~\cite{diffusion, ho2022classifier, balaji2022ediff} primarily operated directly in pixel space, aiming to capture the distributions images in a single stage. However, this approach proved both challenging and inefficient for high-resolution image generation, leading to the development of cascaded models~\cite{saharia2022image, gu2023matryoshka, ho2022cascaded, kim2024pagoda} that generate images through a sequence of stages. These cascaded models typically begin with the generation of a low-resolution image, which is subsequently upscaled by super-resolution models to achieve higher resolutions. However, the diffusion-based super-resolution process often requires starting from pure noise, conditioned on lower-resolution outputs, resulting in a time-consuming and inefficient generation process. Additionally, training these models in isolated stages hinders end-to-end optimization and necessitates carefully designed strategies to ensure the super-resolution stages.

Furthermore, recent advancements in pixel-space generation have introduced innovative architectures. Simple Diffusion~\cite{hoogeboom2023simple, hoogeboom2024simpler} proposes a streamlined diffusion framework for high-resolution image synthesis, achieving strong performance on ImageNet through adjustments of model architecture and noise schedules. FractalGen~\cite{li2025fractal} constructs fractal generative models by recursively invoking atomic generative modules, resulting in self-similar architectures that demonstrate strong performance in pixel-by-pixel image generation. TarFlow~\cite{zhai2024normalizing} presents a Transformer-based normalizing flow architecture capable of directly modeling and generating pixels.

\section{PixelFlow}

\subsection{Preliminary: Flow Matching}\label{sec:flow-preliminary}
The Flow Matching algorithm~\cite{albergo2023building, lipman2023flow, liu2023flow} progressively transforms a sample from a prior distribution, which is typically a standard normal distribution, to the target data distribution. This is accomplished by defining a forward process consisting of a sequence of linear paths that directly connect samples from the prior distribution to corresponding samples in the target distribution.
During training, a training example is constructed by first sampling a target sample $\mathbf{x}_1$, drawing noise $\mathbf{x}_0 \sim \mathcal{N}(0, 1)$ from the standard normal distribution, and selecting a timestep $t \in [0, 1]$. The training example is then defined through a linear interpolation:
\begin{align}\label{eq:flow-linear}
    \mathbf{x}_t = t \cdot \mathbf{x}_1 + (1 - t) \cdot \mathbf{x}_0
\end{align}

The model is trained to approximate the velocity defined by an ordinary differential equation (ODE), $\mathbf{v}_t = \frac{d\mathbf{x}_t}{dt}$, enabling it to effectively guide the transformation from the intermediate sample $\mathbf{x}_t$ to the real data sample $\mathbf{x}_1$.

A notable advantage of Flow Matching is its ability to interpolate between two arbitrary distributions, not restricted to using only a standard Gaussian as the source domain. Consequently, in image generation tasks, Flow Matching extends beyond noise-to-image scenarios and can be effectively employed for diverse applications such as image-to-image translation.

\begin{figure}[t]
    \centering
    \includegraphics[width=1\linewidth]{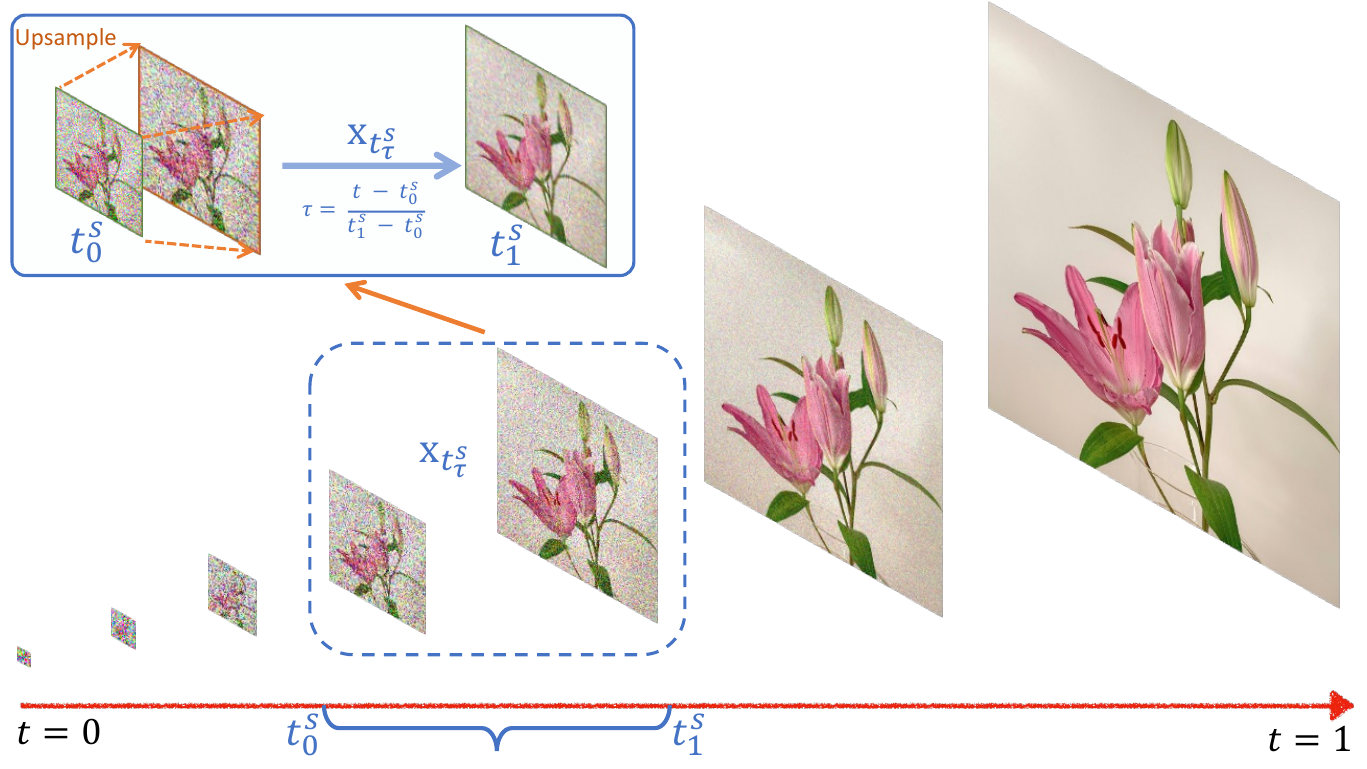}
    \caption{\textbf{PixelFlow for cascaded image generation from pixel space.} We partition the entire generation procedure into series resolution stages. At the beginning of each resolution stage, we upscale the relatively noisy results from the preceding stage and use them as the starting point for the current stage. Consequently, as the resolution enhances, more refined samples can be obtained.}
    \label{fig:pixelflow_pipeline}
    \vspace{9pt}
\end{figure}

\subsection{Multi-Scale Generation in Pixel Space}

PixelFlow generates images by progressively increasing their resolution through a multistage denoising process. To enable this, we construct a multi-scale representation of the target image $\mathbf{x}_1$ by recursively downsampling it by a factor of 2 at each scale. 
As illustrated in \Cref{fig:pixelflow_pipeline}, \ours divides the image generation process into $S$ stages. Each stage $s \in {0, 1, ..., S-1}$ operates over a time interval defined by the start and end states $(\mathbf{x}{t_0^s}, \mathbf{x}{t_1^s})$. In the degenerate case where $S=1$, \ours reduces to a standard single-stage flow matching approach for image generation, similar to recent works~\cite{ma2024sit, esser2024scaling}, but crucially operates in pixel space rather than latent space.

For each stage $s$, we define the starting and ending states as follows:
\begin{align}
    \text{\small Start:} \quad \mathbf{x}_{t_0^s} &= t_0^s \cdot  \textsf{\footnotesize Up}(\textsf{\footnotesize Down}(\mathbf{x}_1, 2^{s+1}) ) + (1 - t_0^s) \cdot \epsilon \\
    \text{\small End:} \quad  \mathbf{x}_{t_1^s} &= t_1^s \cdot  \textsf{\footnotesize Down}(\mathbf{x}_1, 2^s) + (1 - t_1^s) \cdot \epsilon,
\end{align}
where \textsf{\footnotesize Down}($\cdot$) and \textsf{\footnotesize Up}($\cdot$) denote the downsampling and upsampling operations, respectively. Unless otherwise stated, we adopt \texttt{bilinear} interpolation for downsampling and \texttt{nearest} neighbor for upsampling.

To train the model, we sample intermediate representations by linearly interpolating between the start and end states:
\begin{align}\label{eq:train-sample}
    \mathbf{x}_{t_{\tau}^s} = \tau \cdot \mathbf{x}_{t_1^s} + (1 - \tau) \cdot \mathbf{x}_{t_0^s},
\end{align}
where $\tau = \frac{t - t_0^s}{t_1^s - t_0^s}$ is the rescaled timestep~\cite{yan2024perflow, jin2024pyramidal} within the $s$-th stage.

Then our objective is to train a model $\mu_\theta(\cdot)$ to predict the velocity $\mu_\theta(\mathbf{x}_{t_\tau^s, \tau})$ with target as $\mathbf{v}_t = \mathbf{x}_{t_1^s} - \mathbf{x}_{t_0^s}$. We use the mean squared error~(MSE) loss, formally represented as:
\begin{align}
    \mathbb{E}_{s, t, (\mathbf{x}_{t_1^s}, \mathbf{x}_{t_1^s})} || \mu_\theta(\mathbf{x}_{t_\tau^s, \tau}) - \mathbf{v}_t  ||^2
\end{align}

\subsection{Model Architecture}

We instantiate $\mu_\theta(\cdot)$ using a Transformer-based architecture~\cite{vaswani2017attention}, chosen for its simplicity, scalability, and effectiveness in generative modeling. Specifically, our implementation is based on the standard Diffusion Transformer (DiT)~\cite{peebles2023scalable}, employing XL-scale configurations across all experiments. To better align with the PixelFlow framework, we introduce several modifications, as detailed below.

\paragraph{Patchify.} Following the Vision Transformer (ViT) design~\cite{dosovitskiy2020image, peebles2023scalable}, the first layer of PixelFlow is a patch embedding layer, which converts the spatial representation of the input image into a 1D sequence of tokens via a linear projection. In contrast to prior latent transformers~\cite{peebles2023scalable, ma2024sit, esser2024scaling} that operate on VAE-encoded latents, PixelFlow directly tokenizes raw pixel inputs.
To support efficient attention across multiple resolutions within a batch, we apply a sequence packing strategy~\cite{dehghani2024patch}, concatenating flattened token sequences of varying lengths—corresponding to different resolutions—along the sequence dimension.

\paragraph{RoPE.} After patchfying, we replace the original sincos positional encoding~\cite{peebles2023scalable} with RoPE~\cite{su2024roformer} to better handle varying image resolutions. RoPE has shown strong performance in enabling length extrapolation, particularly in large language models. To adapt it for 2D image data, we apply 2D-RoPE by independently applying 1D-RoPE to the height and width dimensions, with each dimension occupying half of the hidden state.

\begin{figure*}[t]
    \centering
    \includegraphics[width=1\linewidth]{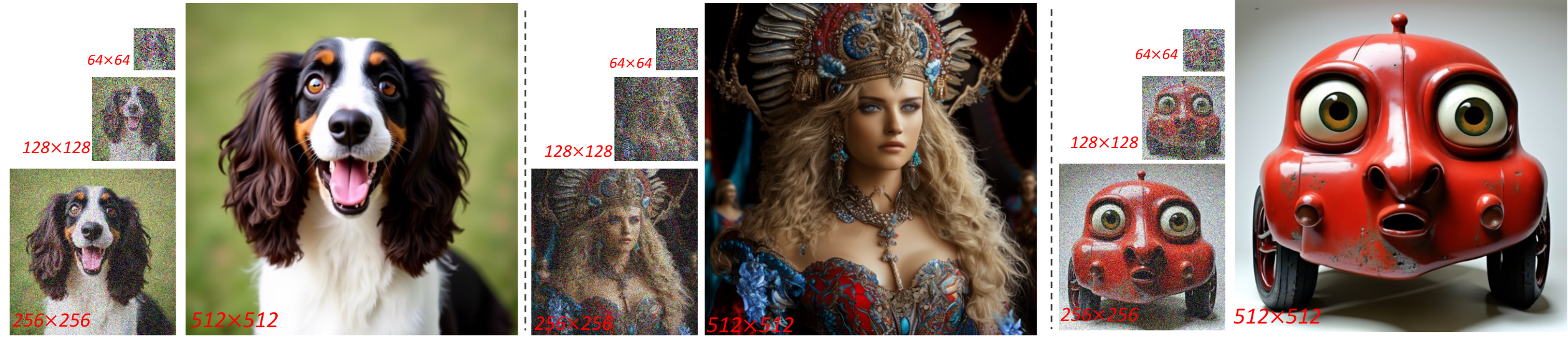}
    \caption{\textbf{Visualization of intermediate result of cascaded stages.} We extract the intermediate results from each of the four stages for direct visualization. We observed a clear denoising process at various resolution stages.
    }\label{fig:stage-demo}
\end{figure*}

\paragraph{Resolution Embedding.} Since PixelFlow operates across multiple resolutions using a shared set of model parameters, we introduce an additional \emph{resolution embedding} to distinguish between resolutions. Specifically, we use the absolute resolution of the feature map after patch embedding as a conditional signal. This signal is encoded using sinusoidal position embedding~\cite{vaswani2017attention} and added to the timestep embedding before being passed into the model.

\paragraph{Text-to-Image Generation.} 
While class-conditional image generation typically integrates conditioning information through adaptive layer normalization (adaLN)\cite{peebles2023scalable}, we extend PixelFlow to support text-to-image generation by introducing a cross-attention layer after each self-attention layer within every Transformer block~\cite{chen2024gentron, chen2023pixart}. This design allows the model to effectively align visual features with the textual input at every stage of the generation process. Following recent work~\cite{sun2024autoregressive, chen2025goku}, we adopt the Flan-T5-XL language model~\cite{chung2024scaling} to extract rich text embeddings, which serve as conditioning signals throughout the network.

\subsection{Training and Inference}
To facilitate efficient training, we uniformly sample training examples from all resolution stages using the interpolation scheme defined in \Cref{eq:train-sample}. Additionally, we employ the sequence packing technique~\cite{dehghani2024patch}, which enables joint training of scale-variant examples within a single mini-batch, improving both efficiency and scalability.

During inference, the generation process begins with pure Gaussian noise at the lowest resolution and progressively transitions to higher resolutions through multiple stages. Within each resolution stage, we apply standard flow-based sampling, using either the Euler discrete sampler~\cite{esser2024scaling} or the Dopri5 solver, depending on the desired trade-off between speed and accuracy. To ensure smooth and coherent transitions across scales, we adopt an renoising strategy~\cite{teng2023relay, jin2024pyramidal}, which effectively mitigates the \emph{jumping point} issue~\cite{campbell2023transdimensional} often observed in multi-scale generation pipelines.

\section{Experiments}
In this section, we first detail our experimental setup in \cref{sec:exp-setup}. Subsequently, we analyze key components of our approach, including model design (\cref{sec:exp-model-design}) and inference configurations (\cref{sec:exp-inference}). Finally, we benchmark \ours against state-of-the-art methods on class-~(\cref{sec:exp-c2i}) and text-to-image~(\cref{sec:exp-t2i}) generation tasks.

\subsection{Experimental Setup}\label{sec:exp-setup}
We evaluate \ours for class-conditional image generation on the ImageNet-1K~\cite{deng2009imagenet} dataset. Unless stated otherwise, we train \ours at 256$\times$256 resolution.
All models are trained using the AdamW optimizer~\cite{kingma2014adam, loshchilov2018decoupled} with a constant learning rate of $1\times10^{-4}$. 
Performance is primarily measured by Fr\'echet Inception Distance~(FID) using the standard evaluation toolkit\footnote{\url{https://github.com/openai/guided-diffusion}}. We also report Inception Score~(IS)~\cite{salimans2016improved}, sFID~\cite{nash2021generating}, and Precision/Recall~\cite{kynkaanniemi2019improved}.

For text-conditional image generation, we progressively train \ours from 256$\times$256 up to 1024$\times$1024 resolution. We include qualitative comparisons with current start-of-the-art generative models, along with quantitative assessments on popular benchmarks such as T2I-CompBench~\cite{huang2023t2i-compbench}, GenEval~\cite{ghosh2024geneval}, and DPG-Bench~\cite{hu2024ella_dbgbench}.

\subsection{Model Design}\label{sec:exp-model-design}

\paragraph{Kickoff sequence length.} 
In principle, \ours can be trained to progressively increase resolution from very low resolution (\eg, $1\times1$) up to the target resolution. However, this approach is inefficient in practice, as tokens at extremely low resolutions convey limited meaningful information. Furthermore, allocating excessive timesteps to very short sequences underutilizes the computational capacity of modern GPUs, resulting in decreased model FLOPS utilizationt. Therefore, we explore how varying the resolution at which image generation begins, which we call \emph{kickoff image resolution}, impacts overall performance.

\begin{table}[ht]
    \tablestyle{3pt}{1}
    \small
    \begin{tabular}{c c c c c c c}
    \toprule
    kickoff seq. len.  & FID $\downarrow$ & sFID $\downarrow$ & IS $\uparrow$ & Precision $\uparrow$ & Recall $\uparrow$ \\
    \midrule
    32$\times$32  & 3.34 & 6.11 & 84.75 & 0.78 & 0.57 \\
    8$\times$8  & 3.21 & 6.23 & 78.50 & 0.78 & 0.56 \\
    2$\times$2  & 3.49 & 6.45 & 67.81 & 0.78 & 0.54 \\
    \bottomrule
    \end{tabular}
    \caption{\textbf{Effect of kickoff sequence length.}  All models are trained with 600k iterations on ImageNet-1K. Patch size is 2$\times$2 and target image resolution is 64$\times$64.}
    \label{tab:ablation-kickoff-seq}
\end{table}

For our transformer-based backbone, the number of tokens involved in attention operations is determined by the raw image resolution and the patch size. In this experiment, we maintain a consistent patch size of 2$\times$2~\cite{peebles2023scalable}, making the kickoff sequence length directly dependent on the kickoff image resolution. Specifically, we evaluate three kickoff sequence length—2$\times$2, 8$\times$8, and 32$\times$32—while keeping the target resolution fixed at 64$\times$64. Notably, the 32$\times$32 setting represents a vanilla pixel-based approach without cascading across resolutions. 

As shown in \Cref{tab:ablation-kickoff-seq}, among these configurations, the 8$\times$8 kickoff sequence length achieves comparable or even slightly improved FID compared to the 32$\times$32 baseline. This suggests that initiating generation from an appropriately smaller resolution and progressively scaling up can maintain generation quality while improving computational efficiency by allocating fewer computations to the largest resolution stage. Conversely, reducing the kickoff sequence length further to 2$\times$2 results in a performance degradation, likely because tokens at extremely low resolutions provide limited useful information and insufficient guidance for subsequent generation steps. Taking into account both generation quality and computational efficiency, we therefore adopt 8$\times$8 as our default kickoff sequence length.

\begin{table}[!t]
    \tablestyle{2pt}{1}
    \small
    \begin{tabular}{c c  c c c c c }
    \toprule
    patch size  & FID $\downarrow$ & sFID $\downarrow$ & IS $\uparrow$ & Precision $\uparrow$ & Recall $\uparrow$  & speed$^\dagger$ \\
    \midrule
    \multicolumn{7}{c}{\emph{target res. 64$\times$64; kickoff seq. len. 2$\times$2; 600K iters}} \\
    \midrule
    2$\times$2 & 3.49 & 6.45 & 67.81 & 0.78 & 0.54 &  1.28 \\
    4$\times$4 & 3.41 & 5.52 & 68.83 & 0.77 & 0.56 &  0.58 \\
    \midrule
    \multicolumn{7}{c}{\emph{target res. 256$\times$256; kickoff seq. len. 2$\times$2; 100K iters}} \\
    \midrule
    2$\times$2 & 28.50 & 6.40 & 47.37 & 0.58 & 0.53 & 30.88 \\
    4$\times$4 & 33.17 & 7.71 & 42.29 & 0.57 & 0.52 & 7.31 \\
    8$\times$8 & 47.50 & 9.63 & 31.19 & 0.45 & 0.50 & 3.96 \\
    \midrule
    \multicolumn{7}{c}{\emph{target res. 256$\times$256; kickoff seq. len. 2$\times$2; 1600K iters; EMA}} \\
    \midrule
    4$\times$4 & 2.81 & 5.48 & 251.79 & 0.82 & 0.55 & 7.31 \\
    8$\times$8 & 4.65 & 5.42 & 195.50 & 0.79 & 0.54 & 3.96 \\
    \bottomrule
    \end{tabular}
    \caption{\textbf{Effect of patch size}. All models have a kickoff sequence length of 2$\times$2. \textbf{Upper:}  target resolution of 64$\times$64; \textbf{Middle:} target resolution of 256$\times$256 resolution, training with 100K iterations due to computational constraints of patch size 2$\times$2; \textbf{Bottom:} Extended training to 1600K iterations at 256$\times$256 resolution. $^\dagger$Speed measured as number of seconds per sample on a single GPU with a batchsize of 50.}
    \label{tab:ablation-patch-size}
\end{table}

\paragraph{Patch size.} Next, we investigate the impact of patch size on model performance while maintaining a kickoff sequence length of 2$\times$2. Initially, we experiment with a target resolution of 64$\times$64 and compare two patch sizes—2$\times$2 and 4$\times$4—with results presented in the upper section of \Cref{tab:ablation-patch-size}. We observe that PixelFlow achieves very similar performance across these two settings, with the 4$\times$4 patch slightly outperforming the 2$\times$2 patch on four out of five evaluation metrics. Furthermore, using a patch size of 4$\times$4 eliminates the highest-resolution stage required by the 2$\times$2 patch size configuration, thus improving efficiency.

When scaling to a larger target resolution (\ie, 256$\times$256), employing a patch size of 2$\times$2 becomes computationally infeasible due to substantial resource demands, limiting our experiments to only 100K training iterations (middle section of \Cref{tab:ablation-patch-size}). This constraint necessitates adopting larger patch sizes. Although increasing the patch size further to 8$\times$8 significantly enhances computational efficiency, it leads to a noticeable drop in performance quality. Moreover, this performance gap persists even after extended training (1600K iterations), as shown in the bottom section of \Cref{tab:ablation-patch-size}. Considering both generation quality and computational cost, we therefore select a patch size of 4$\times$4 as our default setting.

\begin{table}[t]
\centering
\small
    \begin{subtable}[t]{0.48\textwidth}
    \centering
    \begin{tabular*}{\linewidth}{@{\extracolsep{\fill}}c c c c c c}
    \toprule
    step & FID $\downarrow$ & sFID $\downarrow$ & IS $\uparrow$ & Precision $\uparrow$ & Recall $\uparrow$ \\
    \midrule
    10 & 3.39 & 5.98 & 255.27 & 0.80 & 0.54 \\
    20 & 2.53 & 5.53 & 272.13 & 0.82 & 0.56 \\
    30 & 2.51 & 5.82 & 274.92 & 0.82 & 0.56 \\
    40 & 2.55 & 6.58 & 272.68 & 0.81 & 0.56 \\
    \bottomrule
    \end{tabular*}
    \caption{Effect of number of steps per stage. CFG is a global constant value 1.50, sample function is Euler.}
    \label{tab:infere_step}
    \end{subtable}
    \vspace{2pt}
    
    \begin{subtable}[t]{0.48\textwidth}
    \centering
    \begin{tabular*}{\linewidth}{@{\extracolsep{\fill}}c c c c c c}
    \toprule
    solver & FID $\downarrow$ & sFID $\downarrow$ & IS $\uparrow$ & Precision $\uparrow$ & Recall $\uparrow$ \\
    \midrule
    Euler & 2.51 & 5.82 & 274.92 & 0.82 & 0.56 \\
    Dopri5 & 2.43 & 5.38 & 282.20 & 0.83 & 0.56 \\
    \bottomrule
    \end{tabular*}
    \caption{Effect of sample function. CFG is a global constant value 1.50, the number of steps per stage is 30 in Euler, the absolute tolerance is 1e-6 in Dopri5.}
    \label{tab:infere_ode}
    \end{subtable}
    \vspace{2pt}
    
    \begin{subtable}[t]{0.48\textwidth}
    \centering
    \begin{tabular*}{\linewidth}{@{\extracolsep{\fill}}c c c c}
    \toprule
    cfg schedule & cfg max value  & FID $\downarrow$  & IS $\uparrow$ \\
    \midrule
    global constant & 1.50 & 2.43 & 282.2 \\
    stage-wise constant & 2.40 & 1.98 & 282.1  \\
    \bottomrule
    \end{tabular*}
    \caption{Effect of classifier-free guidance (CFG) setting. Sample function is Dopri5 with absolute tolerance 1e-6.}
    \label{tab:infere_cfg}
    \end{subtable}
    \vspace{2pt}
    \caption{\textbf{Inference Setting.} 
The best performance is obtained by CFG step-wise constant with maximum value 2.40 and Dopri5 sample function.}\label{tab:inference}
\end{table}

\begin{figure*}[t]
    \centering
    \includegraphics[width=1\linewidth]{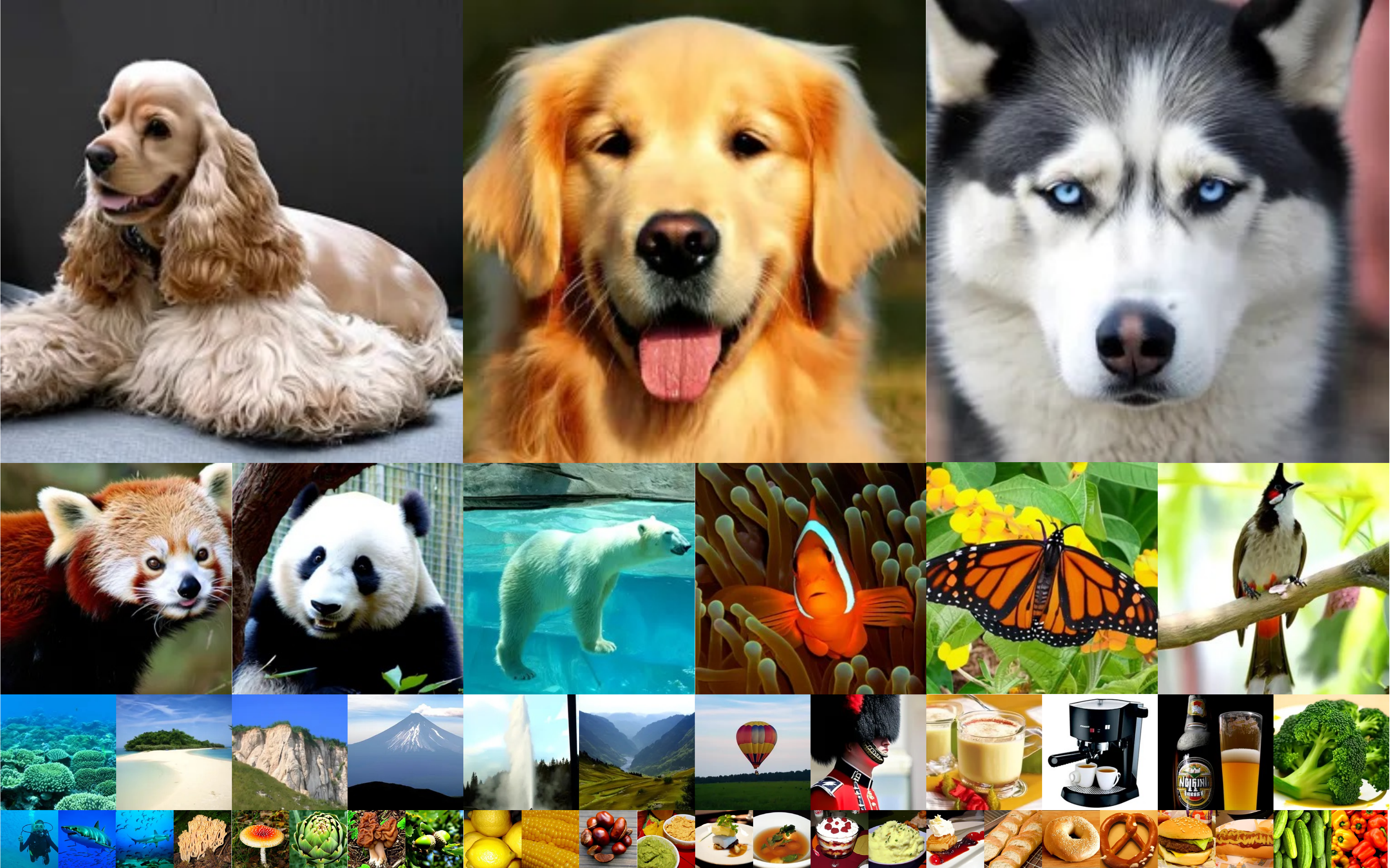}
    \caption{\textbf{Qualitative results} of class-conditional image generation of \ours. All images are 256$\times$256 resolution.}
    \label{fig:c2i_vis}
    \vspace{-2pt}
\end{figure*}

\subsection{Inference Schedule}
\label{sec:exp-inference}
In \Cref{tab:inference}, we provide a detailed analysis of the inference configuration space, including the number of inference steps at each resolution stage, the choice of ODE solver, and the scheduling of classifier-free guidance (CFG).

\paragraph{Number of sample steps.} 
In \Cref{tab:infere_step}, we evaluate the impact of the number of inference steps per resolution stage on generation quality. As the number of steps increases, we observe consistent improvements in FID, sFID, and IS, with the best overall performance achieved at 30 steps. Beyond this point, gains saturate and even slightly decline, indicating diminishing returns.

A notable advantage of \ours is its flexibility in assigning different numbers of sampling steps to each resolution stage during inference. This adaptive configuration allows fine-grained control over the sampling process, enabling performance–efficiency trade-offs. Moving beyond a uniform setting and exploring more granular stage-specific step allocations holds the potential for further performance enhancements.

\paragraph{ODE Solver.} We further investigate the effect of the ODE solver type on generation quality. As shown in \Cref{tab:infere_ode}, we compare the first-order Euler solver with the adaptive higher-order Dormand–Prince~(Dopri5) solver~\cite{dormand1980family}. The results indicate that Dopri5 consistently outperforms Euler across most evaluation metrics, achieving lower FID and sFID scores, a higher Inception Score, and slightly better precision, while maintaining similar recall. This demonstrates that more accurate and adaptive solvers, such as Dopri5, can better capture the generative dynamics, leading to higher-quality samples—though often with increased computational cost.

\paragraph{CFG Schedule.} Inspired by the recent process~\cite{maskgit, cfg_limited, cfg_analysis}, we propose a stage-wise CFG schedule, where different stages apply different CFG values, and from the early stage to the later stage, the value increases from 1 to $\text{CFG}_{\text{max}}$. In the condition of 4 stages, we find that 0, 1/6, 2/3 and 1 of the $(\text{CFG}_{\text{max}} - 1)$  give the best FID performance. The comparison between global constant CFG and stage-wise CFG is shown in \Cref{tab:infere_cfg}, in which we search the best CFG value for each method. Our proposed stage-wise CFG boosts the FID performance from 2.43 to 1.98.

\begin{table}[t]
    \centering 
    \tablestyle{2pt}{1}
    \small
    \begin{tabular}{l c c c c c}
    \toprule
    Model & FID $\downarrow$ & sFID $\downarrow$ & IS $\uparrow$ & Precision $\uparrow$ & Recall $\uparrow$ \\
    \midrule
    \multicolumn{6}{c}{\emph{Latent Space}} \\
    \midrule
    LDM-4-G \cite{rombach2022high} & 3.60 & - &247.7 & 0.87 &0.48 \\
    DiT-XL/2 \cite{peebles2023scalable} & 2.27 & 4.60 & 278.2 & 0.83 & 0.57 \\
    SiT-XL/2 \cite{ma2024sit} & 2.06 & 4.49 & 277.5 & 0.83 & 0.59 \\

    \midrule
    \multicolumn{6}{c}{\emph{Pixel Space}} \\
    \midrule
    ADM-G \cite{dhariwal2021diffusion} & 4.59 & 5.25 & 186.7 & 0.82 & 0.52 \\
    ADM-U \cite{dhariwal2021diffusion} & 3.94 & 6.14 & 215.8 & 0.83 & 0.53\\
    CDM \cite{ho2022cascaded} & 4.88 &-& 158.7 &- &- \\
    RIN \cite{jabri2022scalable,chen2023importance} & 3.42 & - & 182.0 &-& - \\
    SD, U-ViT-L \cite{hoogeboom2023simple} & 2.77 &- &211.8 &- &- \\
    MDM \cite{gu2023matryoshka} & 3.51 &- &- & - &- \\
    StyleGAN-XL \cite{sauer2022stylegan} & 2.30 & 4.02 & 265.1 & 0.78 & 0.53 \\
    VDM++ \cite{kingma2024understanding} & 2.12 & - & 267.7 & - & - \\
    PaGoDA \cite{kim2024pagoda} & 1.56 & - & 259.6 &  - & 0.59  \\
    SiD2~\cite{hoogeboom2024simpler} & 1.38 & - &- & - & - \\
    JetFormer~\cite{tschannen2024jetformer} & 6.64 & - & - & 0.69 & 0.56\\

    FractalMAR-H~\cite{li2025fractal} & 6.15  & - & 348.9 & 0.81 & 0.46 \\
    \midrule
    PixelFlow (ours) & 1.98 & 5.83 & 282.1 & 0.81 & 0.60 \\
    \bottomrule
    \end{tabular}
   \caption{\textbf{Comparisons on class-conditional image generation on ImageNet 256$\times$256}. \ours achieves competitive performance compared with latent space based models.}
    \label{tab:fid256x256}
    % \vspace{-8pt}
\end{table}

\vspace{10pt}
\subsection{Comparison on ImageNet Benchmark}\label{sec:exp-c2i}

In \Cref{tab:fid256x256}, we compare \ours with both latent-based and pixel-based image generation models on the ImageNet 256$\times$256 benchmark. \ours achieves an FID of 1.98, representing highly competitive performance relative to state-of-the-art latent-space methods. For instance, it outperforms LDM~\cite{rombach2022high} (FID 3.60), DiT~\cite{peebles2023scalable} (FID 2.27), and SiT~\cite{ma2024sit} (FID 2.06), while achieving comparable IS and recall scores. These results highlight the effectiveness of our design, suggesting that \ours can serve as a strong prototype for high-quality visual generation systems.

Compared with recent pixel-based models, \ours achieves superior sample quality. It notably outperforms FractalMAR-H~\cite{li2025fractal}, and also delivers competitive or better results than strong baselines like ADM-U~\cite{dhariwal2021diffusion}, SiD2~\cite{hoogeboom2024simpler}, and VDM++~\cite{kingma2024understanding}.

We visualize class-conditional image generation of \ours at 256$\times$256 resolution in \Cref{fig:c2i_vis}. We can observe our model is able to generate images of high visual quality across a wide range of classes.

\vspace{6pt}
\subsection{Text-to-Image Generation}\label{sec:exp-t2i}

\paragraph{Settings.}

We adopt a two-stage training strategy for text-to-image generation of \ours. First, the model is initialized with an ImageNet-pretrained checkpoint at a resolution of 256$\times$256 and trained on a subset of the LAION dataset~\citep{schuhmann2022laion} at the same resolution. In the second stage, we fine-tune the model on a curated set of high-aesthetic-quality images at a higher resolution of 512$\times$512. All reported results for \ours are based on this final 512$\times$512 resolution model.

\begin{table}[t]
    \centering
    \tablestyle{3.3pt}{1.1}
    \small
        \begin{tabular}{l|c|ccc|c}
            \toprule
            \multirow{2}{*}{Method}  & \multicolumn{1}{c|}{\footnotesize GenEval}  &  \multicolumn{3}{c|}{T2I-CompBench}  &  DPG  \\
              & {\footnotesize Overall} &  Color  & Shape & Texture &  Bench   \\
            \midrule
            SDv1.5~\citep{rombach2022high}       & 0.43 & 0.3730 & 0.3646 & 0.4219 & 63.18   \\
            DALL-E 2~\citep{ramesh2022hierarchical} & 0.52 & 0.5750 & 0.5464 & 0.6374 & -   \\
            SDv2.1~\citep{rombach2022high}          & 0.50 & 0.5694 & 0.4495 &  0.4982 & -   \\
            SDXL~\citep{podell2023sdxl}             & 0.55 & 0.6369 & 0.5408 & 0.5637 & 74.65 \\
            PixArt-$\alpha$~\citep{chen2023pixart} & 0.48 & 0.6886 & 0.5582 & 0.7044 & 71.11   \\
            DALL-E 3~\citep{betker2023improving}       &~~0.67$^\dagger$  & 0.8110$^\dagger$ & 0.6750$^\dagger$ & 0.8070$^\dagger$ & 83.50$^\dagger$    \\ 
            GenTron~\citep{chen2024gentron} &  -  & 0.7674 & 0.5700 & 0.7150  & - \\
            SD3~\citep{esser2024scaling} & 0.74 & - & - & - & -  \\
            Transfusion~\citep{zhou2024transfusion}  & 0.63 & - & - & - &  - \\
            LlamaGen~\citep{sun2024autoregressive}  & 0.32  & - & - & - & - \\
            Emu 3~\citep{wang2024emu3}     &0.66$^\dagger$ & 0.7913$^\dagger$ & 0.5846$^\dagger$ & 0.7422$^\dagger$ & 80.60  \\
            \midrule            
             \multirow{2}{*}{\ours (ours)}  & 0.60 & 0.7578 & 0.4529 & 0.6006 &  \multirow{2}{*}{77.93}  \\
              &  0.64$^\dagger$ & 0.7689$^\dagger$ & 0.5059$^\dagger$ & 0.6273$^\dagger$ &    \\
            \bottomrule
        \end{tabular}
    \caption{\textbf{Comparison with state-of-the-art models on text-to-image generation benchmarks.} We evaluate on GenEval~\cite{ghosh2024geneval}, T2I-CompBench~\cite{huang2023t2i-compbench} and DPG-Bench~\cite{hu2024ella_dbgbench}. We use $\dagger$ to indicate the result with prompt rewriting.}\label{tab:text2image_evaluation}
\end{table}

\begin{figure*}[!ht]
    \centering
    \includegraphics[width=1.00\linewidth]{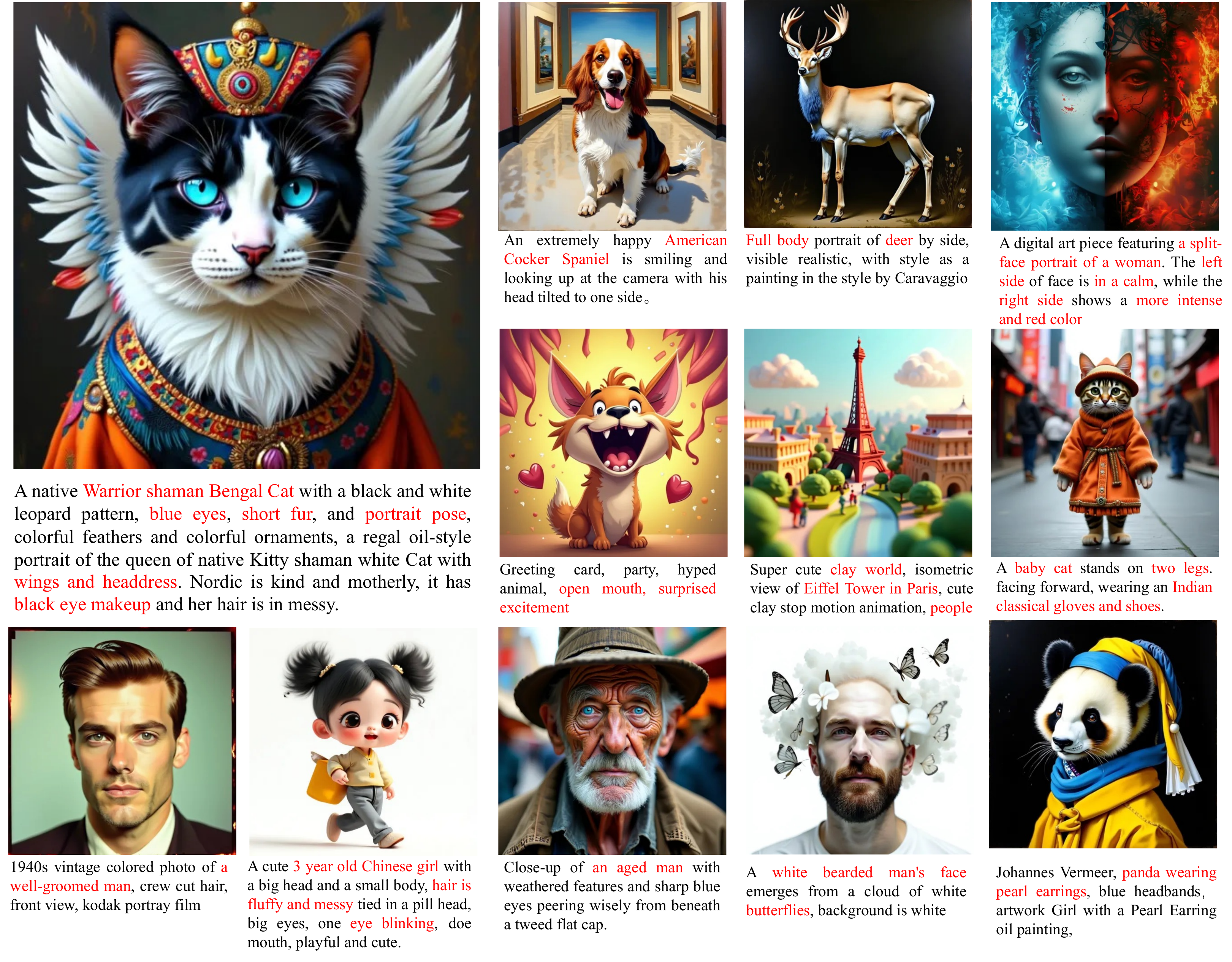}
    \caption{\textbf{Qualitative results} of text-conditional generation of \ours. All images are \textbf{512$\times$512 resolution}. Key components of the prompt are highlighted in \textcolor{red}{\textbf{RED}}.}
    \label{fig:t2i_512}
\end{figure*}

To comprehensively evaluate the performance of \ours-T2I in text-to-image generation, we employ three widely recognized benchmarks, each targeting a different facet of compositional understanding:
\textbf{T2I-CompBench}~\cite{huang2023t2i-compbench} assesses alignment between generated images and complex semantic relationships in text. We evaluate three tasks—color, shape, and texture binding—by generating five images per prompt across 300 prompts per sub-task. Alignment is measured using BLIP-VQA\citep{li2022blip}; \textbf{GenEval}~\cite{ghosh2024geneval} evaluates compositional aspects such as coherence and spatial arrangement. We generate over 2,000 images from 553 prompts and report the average performance across tasks; \textbf{DPG-Bench}~\cite{hu2024ella_dbgbench} focuses on complex textual descriptions, with 4,000 images generated from 1,065 prompts and results averaged across tasks.

\paragraph{Quantitative results.}
As shown in \Cref{tab:text2image_evaluation}, \ours achieves competitive performance across all benchmarks, demonstrating strong compositional understanding in free-form text-to-image generation. It performs particularly well on T2I-CompBench, with high scores in color and texture binding, and solid results on GenEval (0.64) and DPG-Bench (77.93), surpassing many established models. These results underscore \ours as a promising direction for pixel-space image generation conditioned on natural language—showcasing its potential for open-ended, text-driven image synthesis.

\begin{figure*}[t]
\centering
\includegraphics[width=0.80\textwidth]{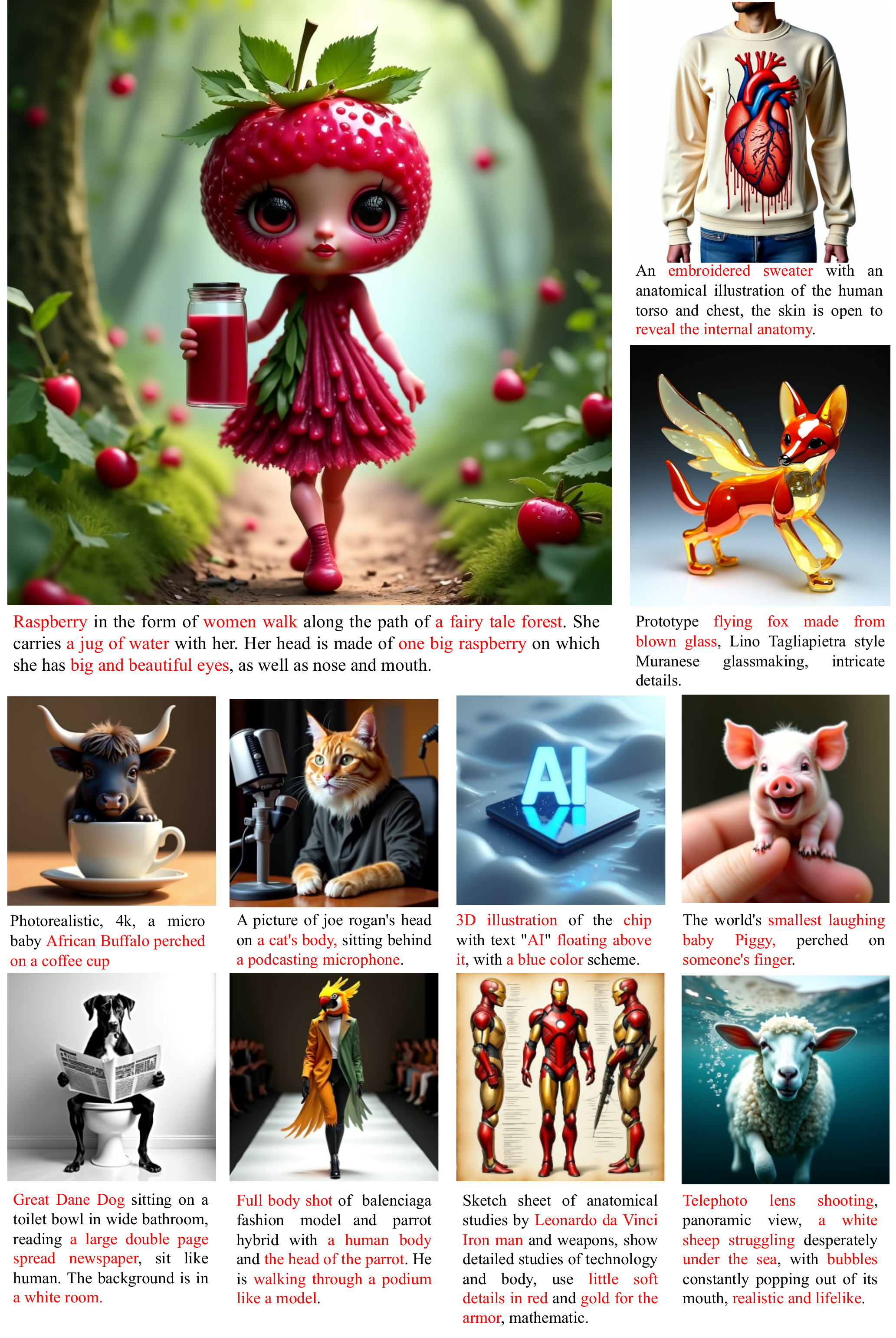}
\caption{\textbf{Qualitative samples of \ours.} We present the generated images of \textbf{1024$\times$1024 resolution}. Key words are highlighted in \textcolor{red}{RED}.}
\label{fig:t2i_1024}
\end{figure*}

\paragraph{Visualization.}
We visualize the intermediate results during the sampling process in \Cref{fig:stage-demo}, specifically showing the final step of each resolution stage. As resolution increases, a clear denoising trend emerges—images become progressively cleaner and less noisy at each stage.
Additional generated samples along with their input text prompts are shown in \Cref{fig:t2i_512} (512$\times$512) and \Cref{fig:t2i_1024} (1024$\times$1024). \ours demonstrates high visual fidelity and strong text-image alignment, effectively capturing key visual elements and their relationships from complex prompts. Notably, it generates fine-grained details—such as animal fur, human hair, and hat textures—highlighting its strong attention to detail in pixel space.

\section{Conclusion}
We introduce PixelFlow, a novel image generation model that re-think the predominance of latent space based models by directly operating on raw pixel space. By directly transforming between different resolution stages, our model exhibits a compelling advantage in simplicity and end-to-end trainability. On both class-conditional image generation and text-to-image generation benchmarks, PixelFlow has been proven to demonstrate competitive image generation capabilities compared to popular latent
space-based methods. We hope that this new perspective will inspire future research in visual generation models. 

\vspace{-5pt}
\paragraph{Limitations} Despite its advantages, \ours still faces certain limitations. Although the model avoids full-resolution computation across all stages, the final stage requires full-resolution attention, which accounts for roughly 80\% of the total inference time. Moreover, we observe that training convergence slows as the sequence length increases. Addressing these challenges presents opportunities for future improvements in efficiency and scalability.

\newpage
{
    \small
    \bibliographystyle{ieeenat_fullname}
    \bibliography{main}

\begin{thebibliography}{70}
\providecommand{\natexlab}[1]{#1}
\providecommand{\url}[1]{\texttt{#1}}
\expandafter\ifx\csname urlstyle\endcsname\relax
  \providecommand{\doi}[1]{doi: #1}\else
  \providecommand{\doi}{doi: \begingroup \urlstyle{rm}\Url}\fi

\bibitem[Albergo and Vanden-Eijnden(2023)]{albergo2023building}
Michael~Samuel Albergo and Eric Vanden-Eijnden.
\newblock Building normalizing flows with stochastic interpolants.
\newblock In \emph{The Eleventh International Conference on Learning Representations}, 2023.

\bibitem[Balaji et~al.(2022)Balaji, Nah, Huang, Vahdat, Song, Zhang, Kreis, Aittala, Aila, Laine, et~al.]{balaji2022ediff}
Yogesh Balaji, Seungjun Nah, Xun Huang, Arash Vahdat, Jiaming Song, Qinsheng Zhang, Karsten Kreis, Miika Aittala, Timo Aila, Samuli Laine, et~al.
\newblock ediff-i: Text-to-image diffusion models with an ensemble of expert denoisers.
\newblock \emph{arXiv preprint arXiv:2211.01324}, 2022.

\bibitem[Betker et~al.(2023)Betker, Goh, Jing, Brooks, Wang, Li, Ouyang, Zhuang, Lee, Guo, et~al.]{betker2023improving}
James Betker, Gabriel Goh, Li Jing, Tim Brooks, Jianfeng Wang, Linjie Li, Long Ouyang, Juntang Zhuang, Joyce Lee, Yufei Guo, et~al.
\newblock Improving image generation with better captions.
\newblock \emph{Computer Science. https://cdn. openai. com/papers/dall-e-3. pdf}, 2\penalty0 (3):\penalty0 8, 2023.

\bibitem[Campbell et~al.(2023)Campbell, Harvey, Weilbach, Bortoli, Rainforth, and Doucet]{campbell2023transdimensional}
Andrew Campbell, William Harvey, Christian~Dietrich Weilbach, Valentin~De Bortoli, Tom Rainforth, and Arnaud Doucet.
\newblock Trans-dimensional generative modeling via jump diffusion models.
\newblock In \emph{Thirty-seventh Conference on Neural Information Processing Systems}, 2023.

\bibitem[Chang et~al.(2022)Chang, Zhang, Jiang, Liu, and Freeman]{maskgit}
Huiwen Chang, Han Zhang, Lu Jiang, Ce Liu, and William~T Freeman.
\newblock Maskgit: Masked generative image transformer.
\newblock In \emph{Proceedings of the IEEE/CVF conference on computer vision and pattern recognition}, pages 11315--11325, 2022.

\bibitem[Chen et~al.(2023)Chen, Yu, Ge, Yao, Xie, Wu, Wang, Kwok, Luo, Lu, et~al.]{chen2023pixart}
Junsong Chen, Jincheng Yu, Chongjian Ge, Lewei Yao, Enze Xie, Yue Wu, Zhongdao Wang, James Kwok, Ping Luo, Huchuan Lu, et~al.
\newblock Pixart-$alpha$: Fast training of diffusion transformer for photorealistic text-to-image synthesis.
\newblock \emph{arXiv preprint arXiv:2310.00426}, 2023.

\bibitem[Chen et~al.(2024)Chen, Xu, Ren, Cong, He, Xie, Sinha, Luo, Xiang, and Perez-Rua]{chen2024gentron}
Shoufa Chen, Mengmeng Xu, Jiawei Ren, Yuren Cong, Sen He, Yanping Xie, Animesh Sinha, Ping Luo, Tao Xiang, and Juan-Manuel Perez-Rua.
\newblock Gentron: Diffusion transformers for image and video generation.
\newblock In \emph{Proceedings of the IEEE/CVF Conference on Computer Vision and Pattern Recognition}, pages 6441--6451, 2024.

\bibitem[Chen et~al.(2025)Chen, Ge, Zhang, Zhang, Zhu, Yang, Hao, Wu, Lai, Hu, et~al.]{chen2025goku}
Shoufa Chen, Chongjian Ge, Yuqi Zhang, Yida Zhang, Fengda Zhu, Hao Yang, Hongxiang Hao, Hui Wu, Zhichao Lai, Yifei Hu, et~al.
\newblock Goku: Flow based video generative foundation models.
\newblock \emph{arXiv preprint arXiv:2502.04896}, 2025.

\bibitem[Chen(2023)]{chen2023importance}
Ting Chen.
\newblock On the importance of noise scheduling for diffusion models.
\newblock \emph{arXiv preprint arXiv:2301.10972}, 2023.

\bibitem[Chung et~al.(2024)Chung, Hou, Longpre, Zoph, Tay, Fedus, Li, Wang, Dehghani, Brahma, et~al.]{chung2024scaling}
Hyung~Won Chung, Le Hou, Shayne Longpre, Barret Zoph, Yi Tay, William Fedus, Yunxuan Li, Xuezhi Wang, Mostafa Dehghani, Siddhartha Brahma, et~al.
\newblock Scaling instruction-finetuned language models.
\newblock \emph{Journal of Machine Learning Research}, 25\penalty0 (70):\penalty0 1--53, 2024.

\bibitem[Dehghani et~al.(2024)Dehghani, Mustafa, Djolonga, Heek, Minderer, Caron, Steiner, Puigcerver, Geirhos, Alabdulmohsin, et~al.]{dehghani2024patch}
Mostafa Dehghani, Basil Mustafa, Josip Djolonga, Jonathan Heek, Matthias Minderer, Mathilde Caron, Andreas Steiner, Joan Puigcerver, Robert Geirhos, Ibrahim~M Alabdulmohsin, et~al.
\newblock Patch n’pack: Navit, a vision transformer for any aspect ratio and resolution.
\newblock \emph{Advances in Neural Information Processing Systems}, 36, 2024.

\bibitem[Deng et~al.(2009)Deng, Dong, Socher, Li, Li, and Fei-Fei]{deng2009imagenet}
Jia Deng, Wei Dong, Richard Socher, Li-Jia Li, Kai Li, and Li Fei-Fei.
\newblock Imagenet: A large-scale hierarchical image database.
\newblock In \emph{2009 IEEE conference on computer vision and pattern recognition}, pages 248--255. Ieee, 2009.

\bibitem[Dhariwal and Nichol(2021)]{dhariwal2021diffusion}
Prafulla Dhariwal and Alexander Nichol.
\newblock Diffusion models beat gans on image synthesis.
\newblock \emph{Advances in neural information processing systems}, 34:\penalty0 8780--8794, 2021.

\bibitem[Dormand and Prince(1980)]{dormand1980family}
John~R Dormand and Peter~J Prince.
\newblock A family of embedded runge-kutta formulae.
\newblock \emph{Journal of computational and applied mathematics}, 6\penalty0 (1):\penalty0 19--26, 1980.

\bibitem[Dosovitskiy(2020)]{dosovitskiy2020image}
Alexey Dosovitskiy.
\newblock An image is worth 16x16 words: Transformers for image recognition at scale.
\newblock \emph{arXiv preprint arXiv:2010.11929}, 2020.

\bibitem[Esser et~al.(2021)Esser, Rombach, and Ommer]{esser2021taming}
Patrick Esser, Robin Rombach, and Bjorn Ommer.
\newblock Taming transformers for high-resolution image synthesis.
\newblock In \emph{Proceedings of the IEEE/CVF conference on computer vision and pattern recognition}, pages 12873--12883, 2021.

\bibitem[Esser et~al.(2024)Esser, Kulal, Blattmann, Entezari, M{\"u}ller, Saini, Levi, Lorenz, Sauer, Boesel, et~al.]{esser2024scaling}
Patrick Esser, Sumith Kulal, Andreas Blattmann, Rahim Entezari, Jonas M{\"u}ller, Harry Saini, Yam Levi, Dominik Lorenz, Axel Sauer, Frederic Boesel, et~al.
\newblock Scaling rectified flow transformers for high-resolution image synthesis.
\newblock In \emph{Forty-first International Conference on Machine Learning}, 2024.

\bibitem[Evans et~al.(2024)Evans, Carr, Taylor, Hawley, and Pons]{evans2024fast}
Zach Evans, CJ Carr, Josiah Taylor, Scott~H Hawley, and Jordi Pons.
\newblock Fast timing-conditioned latent audio diffusion.
\newblock In \emph{Forty-first International Conference on Machine Learning}, 2024.

\bibitem[Ghosh et~al.(2024)Ghosh, Hajishirzi, and Schmidt]{ghosh2024geneval}
Dhruba Ghosh, Hannaneh Hajishirzi, and Ludwig Schmidt.
\newblock Geneval: An object-focused framework for evaluating text-to-image alignment.
\newblock \emph{Advances in Neural Information Processing Systems}, 36, 2024.

\bibitem[Gu et~al.(2023)Gu, Zhai, Zhang, Susskind, and Jaitly]{gu2023matryoshka}
Jiatao Gu, Shuangfei Zhai, Yizhe Zhang, Joshua~M Susskind, and Navdeep Jaitly.
\newblock Matryoshka diffusion models.
\newblock In \emph{The Twelfth International Conference on Learning Representations}, 2023.

\bibitem[Ho and Salimans(2022)]{ho2022classifier}
Jonathan Ho and Tim Salimans.
\newblock Classifier-free diffusion guidance.
\newblock \emph{arXiv preprint arXiv:2207.12598}, 2022.

\bibitem[Ho et~al.(2022)Ho, Saharia, Chan, Fleet, Norouzi, and Salimans]{ho2022cascaded}
Jonathan Ho, Chitwan Saharia, William Chan, David~J Fleet, Mohammad Norouzi, and Tim Salimans.
\newblock Cascaded diffusion models for high fidelity image generation.
\newblock \emph{Journal of Machine Learning Research}, 23\penalty0 (47):\penalty0 1--33, 2022.

\bibitem[Hong et~al.(2022)Hong, Ding, Zheng, Liu, and Tang]{hong2022cogvideo}
Wenyi Hong, Ming Ding, Wendi Zheng, Xinghan Liu, and Jie Tang.
\newblock Cogvideo: Large-scale pretraining for text-to-video generation via transformers.
\newblock \emph{arXiv preprint arXiv:2205.15868}, 2022.

\bibitem[Hoogeboom et~al.(2023)Hoogeboom, Heek, and Salimans]{hoogeboom2023simple}
Emiel Hoogeboom, Jonathan Heek, and Tim Salimans.
\newblock simple diffusion: End-to-end diffusion for high resolution images.
\newblock In \emph{International Conference on Machine Learning}, pages 13213--13232. PMLR, 2023.

\bibitem[Hoogeboom et~al.(2024)Hoogeboom, Mensink, Heek, Lamerigts, Gao, and Salimans]{hoogeboom2024simpler}
Emiel Hoogeboom, Thomas Mensink, Jonathan Heek, Kay Lamerigts, Ruiqi Gao, and Tim Salimans.
\newblock Simpler diffusion (sid2): 1.5 fid on imagenet512 with pixel-space diffusion.
\newblock \emph{arXiv preprint arXiv:2410.19324}, 2024.

\bibitem[Hu et~al.(2024)Hu, Wang, Fang, Fu, Cheng, and Yu]{hu2024ella_dbgbench}
Xiwei Hu, Rui Wang, Yixiao Fang, Bin Fu, Pei Cheng, and Gang Yu.
\newblock Ella: Equip diffusion models with llm for enhanced semantic alignment.
\newblock \emph{arXiv preprint arXiv:2403.05135}, 2024.

\bibitem[Huang et~al.(2023)Huang, Sun, Xie, Li, and Liu]{huang2023t2i-compbench}
Kaiyi Huang, Kaiyue Sun, Enze Xie, Zhenguo Li, and Xihui Liu.
\newblock T2i-compbench: A comprehensive benchmark for open-world compositional text-to-image generation.
\newblock \emph{Advances in Neural Information Processing Systems}, 36:\penalty0 78723--78747, 2023.

\bibitem[Jabri et~al.(2022)Jabri, Fleet, and Chen]{jabri2022scalable}
Allan Jabri, David Fleet, and Ting Chen.
\newblock Scalable adaptive computation for iterative generation.
\newblock \emph{arXiv preprint arXiv:2212.11972}, 2022.

\bibitem[Jin et~al.(2024)Jin, Sun, Li, Xu, Jiang, Zhuang, Huang, Song, Mu, and Lin]{jin2024pyramidal}
Yang Jin, Zhicheng Sun, Ningyuan Li, Kun Xu, Hao Jiang, Nan Zhuang, Quzhe Huang, Yang Song, Yadong Mu, and Zhouchen Lin.
\newblock Pyramidal flow matching for efficient video generative modeling.
\newblock \emph{arXiv preprint arXiv:2410.05954}, 2024.

\bibitem[Kim et~al.(2024)Kim, Lai, Liao, Takida, Murata, Uesaka, Mitsufuji, and Ermon]{kim2024pagoda}
Dongjun Kim, Chieh-Hsin Lai, Wei-Hsiang Liao, Yuhta Takida, Naoki Murata, Toshimitsu Uesaka, Yuki Mitsufuji, and Stefano Ermon.
\newblock Pagoda: Progressive growing of a one-step generator from a low-resolution diffusion teacher.
\newblock \emph{arXiv preprint arXiv:2405.14822}, 2024.

\bibitem[Kingma and Gao(2024)]{kingma2024understanding}
Diederik Kingma and Ruiqi Gao.
\newblock Understanding diffusion objectives as the elbo with simple data augmentation.
\newblock \emph{Advances in Neural Information Processing Systems}, 36, 2024.

\bibitem[Kingma and Ba(2015)]{kingma2014adam}
Diederik~P. Kingma and Jimmy Ba.
\newblock Adam: A method for stochastic optimization.
\newblock In \emph{International Conference on Learning Representations}, 2015.

\bibitem[Kynk{\"a}{\"a}nniemi et~al.(2019)Kynk{\"a}{\"a}nniemi, Karras, Laine, Lehtinen, and Aila]{kynkaanniemi2019improved}
Tuomas Kynk{\"a}{\"a}nniemi, Tero Karras, Samuli Laine, Jaakko Lehtinen, and Timo Aila.
\newblock Improved precision and recall metric for assessing generative models.
\newblock \emph{Advances in neural information processing systems}, 32, 2019.

\bibitem[Kynk{\"a}{\"a}nniemi et~al.(2024)Kynk{\"a}{\"a}nniemi, Aittala, Karras, Laine, Aila, and Lehtinen]{cfg_limited}
Tuomas Kynk{\"a}{\"a}nniemi, Miika Aittala, Tero Karras, Samuli Laine, Timo Aila, and Jaakko Lehtinen.
\newblock Applying guidance in a limited interval improves sample and distribution quality in diffusion models.
\newblock \emph{arXiv preprint arXiv:2404.07724}, 2024.

\bibitem[Labs(2024)]{flux2024}
Black~Forest Labs.
\newblock Flux.
\newblock \url{https://github.com/black-forest-labs/flux}, 2024.

\bibitem[Li et~al.(2022)Li, Li, Xiong, and Hoi]{li2022blip}
Junnan Li, Dongxu Li, Caiming Xiong, and Steven Hoi.
\newblock Blip: Bootstrapping language-image pre-training for unified vision-language understanding and generation.
\newblock In \emph{International conference on machine learning}, pages 12888--12900. PMLR, 2022.

\bibitem[Li et~al.(2025)Li, Sun, Fan, and He]{li2025fractal}
Tianhong Li, Qinyi Sun, Lijie Fan, and Kaiming He.
\newblock Fractal generative models.
\newblock \emph{arXiv preprint arXiv:2502.17437}, 2025.

\bibitem[Lipman et~al.(2023)Lipman, Chen, Ben-Hamu, Nickel, and Le]{lipman2023flow}
Yaron Lipman, Ricky T.~Q. Chen, Heli Ben-Hamu, Maximilian Nickel, and Matthew Le.
\newblock Flow matching for generative modeling.
\newblock In \emph{The Eleventh International Conference on Learning Representations}, 2023.

\bibitem[Liu et~al.(2023{\natexlab{a}})Liu, Chen, Yuan, Mei, Liu, Mandic, Wang, and Plumbley]{liu2023audioldm}
Haohe Liu, Zehua Chen, Yi Yuan, Xinhao Mei, Xubo Liu, Danilo Mandic, Wenwu Wang, and Mark~D Plumbley.
\newblock Audioldm: Text-to-audio generation with latent diffusion models.
\newblock \emph{arXiv preprint arXiv:2301.12503}, 2023{\natexlab{a}}.

\bibitem[Liu et~al.(2023{\natexlab{b}})Liu, Gong, and qiang liu]{liu2023flow}
Xingchao Liu, Chengyue Gong, and qiang liu.
\newblock Flow straight and fast: Learning to generate and transfer data with rectified flow.
\newblock In \emph{The Eleventh International Conference on Learning Representations}, 2023{\natexlab{b}}.

\bibitem[Loshchilov and Hutter(2019)]{loshchilov2018decoupled}
Ilya Loshchilov and Frank Hutter.
\newblock Decoupled weight decay regularization.
\newblock In \emph{International Conference on Learning Representations}, 2019.

\bibitem[Ma et~al.(2024)Ma, Goldstein, Albergo, Boffi, Vanden-Eijnden, and Xie]{ma2024sit}
Nanye Ma, Mark Goldstein, Michael~S Albergo, Nicholas~M Boffi, Eric Vanden-Eijnden, and Saining Xie.
\newblock Sit: Exploring flow and diffusion-based generative models with scalable interpolant transformers.
\newblock \emph{arXiv preprint arXiv:2401.08740}, 2024.

\bibitem[Nash et~al.(2021)Nash, Menick, Dieleman, and Battaglia]{nash2021generating}
Charlie Nash, Jacob Menick, Sander Dieleman, and Peter~W Battaglia.
\newblock Generating images with sparse representations.
\newblock \emph{arXiv preprint arXiv:2103.03841}, 2021.

\bibitem[NVIDIA(2024)]{edifyimage}
NVIDIA.
\newblock Edify image: High-quality image generation with pixel space laplacian diffusion model.
\newblock \emph{arXiv preprint arXiv:2411.07126}, 2024.

\bibitem[Peebles and Xie(2023)]{peebles2023scalable}
William Peebles and Saining Xie.
\newblock Scalable diffusion models with transformers.
\newblock In \emph{Proceedings of the IEEE/CVF International Conference on Computer Vision}, pages 4195--4205, 2023.

\bibitem[Pernias et~al.(2023)Pernias, Rampas, Richter, Pal, and Aubreville]{stablecascade}
Pablo Pernias, Dominic Rampas, Mats~L Richter, Christopher~J Pal, and Marc Aubreville.
\newblock W{\"u}rstchen: An efficient architecture for large-scale text-to-image diffusion models.
\newblock \emph{arXiv preprint arXiv:2306.00637}, 2023.

\bibitem[Podell et~al.(2023)Podell, English, Lacey, Blattmann, Dockhorn, M{\"u}ller, Penna, and Rombach]{podell2023sdxl}
Dustin Podell, Zion English, Kyle Lacey, Andreas Blattmann, Tim Dockhorn, Jonas M{\"u}ller, Joe Penna, and Robin Rombach.
\newblock Sdxl: Improving latent diffusion models for high-resolution image synthesis.
\newblock \emph{arXiv preprint arXiv:2307.01952}, 2023.

\bibitem[Ramesh et~al.(2021)Ramesh, Pavlov, Goh, Gray, Voss, Radford, Chen, and Sutskever]{ramesh2021zero}
Aditya Ramesh, Mikhail Pavlov, Gabriel Goh, Scott Gray, Chelsea Voss, Alec Radford, Mark Chen, and Ilya Sutskever.
\newblock Zero-shot text-to-image generation.
\newblock In \emph{International conference on machine learning}, pages 8821--8831. Pmlr, 2021.

\bibitem[Ramesh et~al.(2022)Ramesh, Dhariwal, Nichol, Chu, and Chen]{ramesh2022hierarchical}
Aditya Ramesh, Prafulla Dhariwal, Alex Nichol, Casey Chu, and Mark Chen.
\newblock Hierarchical text-conditional image generation with clip latents.
\newblock \emph{arXiv preprint arXiv:2204.06125}, 1\penalty0 (2):\penalty0 3, 2022.

\bibitem[Rombach et~al.(2022)Rombach, Blattmann, Lorenz, Esser, and Ommer]{rombach2022high}
Robin Rombach, Andreas Blattmann, Dominik Lorenz, Patrick Esser, and Bj{\"o}rn Ommer.
\newblock High-resolution image synthesis with latent diffusion models.
\newblock In \emph{Proceedings of the IEEE/CVF conference on computer vision and pattern recognition}, pages 10684--10695, 2022.

\bibitem[Saharia et~al.(2022{\natexlab{a}})Saharia, Chan, Saxena, Li, Whang, Denton, Ghasemipour, Gontijo~Lopes, Karagol~Ayan, Salimans, et~al.]{saharia2022photorealistic}
Chitwan Saharia, William Chan, Saurabh Saxena, Lala Li, Jay Whang, Emily~L Denton, Kamyar Ghasemipour, Raphael Gontijo~Lopes, Burcu Karagol~Ayan, Tim Salimans, et~al.
\newblock Photorealistic text-to-image diffusion models with deep language understanding.
\newblock \emph{Advances in neural information processing systems}, 35:\penalty0 36479--36494, 2022{\natexlab{a}}.

\bibitem[Saharia et~al.(2022{\natexlab{b}})Saharia, Ho, Chan, Salimans, Fleet, and Norouzi]{saharia2022image}
Chitwan Saharia, Jonathan Ho, William Chan, Tim Salimans, David~J Fleet, and Mohammad Norouzi.
\newblock Image super-resolution via iterative refinement.
\newblock \emph{IEEE transactions on pattern analysis and machine intelligence}, 45\penalty0 (4):\penalty0 4713--4726, 2022{\natexlab{b}}.

\bibitem[Salimans et~al.(2016)Salimans, Goodfellow, Zaremba, Cheung, Radford, and Chen]{salimans2016improved}
Tim Salimans, Ian Goodfellow, Wojciech Zaremba, Vicki Cheung, Alec Radford, and Xi Chen.
\newblock Improved techniques for training gans.
\newblock \emph{Advances in neural information processing systems}, 29, 2016.

\bibitem[Sauer et~al.(2022)Sauer, Schwarz, and Geiger]{sauer2022stylegan}
Axel Sauer, Katja Schwarz, and Andreas Geiger.
\newblock Stylegan-xl: Scaling stylegan to large diverse datasets.
\newblock In \emph{ACM SIGGRAPH 2022 conference proceedings}, pages 1--10, 2022.

\bibitem[Schuhmann et~al.(2022)Schuhmann, Beaumont, Vencu, Gordon, Wightman, Cherti, Coombes, Katta, Mullis, Wortsman, et~al.]{schuhmann2022laion}
Christoph Schuhmann, Romain Beaumont, Richard Vencu, Cade Gordon, Ross Wightman, Mehdi Cherti, Theo Coombes, Aarush Katta, Clayton Mullis, Mitchell Wortsman, et~al.
\newblock Laion-5b: An open large-scale dataset for training next generation image-text models.
\newblock \emph{Advances in Neural Information Processing Systems}, 35:\penalty0 25278--25294, 2022.

\bibitem[Sohl-Dickstein et~al.(2015)Sohl-Dickstein, Weiss, Maheswaranathan, and Ganguli]{diffusion}
Jascha Sohl-Dickstein, Eric Weiss, Niru Maheswaranathan, and Surya Ganguli.
\newblock Deep unsupervised learning using nonequilibrium thermodynamics.
\newblock In \emph{International conference on machine learning}, pages 2256--2265. PMLR, 2015.

\bibitem[Stan et~al.(2023)Stan, Wofk, Fox, Redden, Saxton, Yu, Aflalo, Tseng, Nonato, Muller, et~al.]{stan2023ldm3d}
Gabriela Ben~Melech Stan, Diana Wofk, Scottie Fox, Alex Redden, Will Saxton, Jean Yu, Estelle Aflalo, Shao-Yen Tseng, Fabio Nonato, Matthias Muller, et~al.
\newblock Ldm3d: Latent diffusion model for 3d.
\newblock \emph{arXiv preprint arXiv:2305.10853}, 2023.

\bibitem[Su et~al.(2024)Su, Ahmed, Lu, Pan, Bo, and Liu]{su2024roformer}
Jianlin Su, Murtadha Ahmed, Yu Lu, Shengfeng Pan, Wen Bo, and Yunfeng Liu.
\newblock Roformer: Enhanced transformer with rotary position embedding.
\newblock \emph{Neurocomputing}, 568:\penalty0 127063, 2024.

\bibitem[Sun et~al.(2024)Sun, Jiang, Chen, Zhang, Peng, Luo, and Yuan]{sun2024autoregressive}
Peize Sun, Yi Jiang, Shoufa Chen, Shilong Zhang, Bingyue Peng, Ping Luo, and Zehuan Yuan.
\newblock Autoregressive model beats diffusion: Llama for scalable image generation.
\newblock \emph{arXiv preprint arXiv:2406.06525}, 2024.

\bibitem[Teng et~al.(2023)Teng, Zheng, Ding, Hong, Wangni, Yang, and Tang]{teng2023relay}
Jiayan Teng, Wendi Zheng, Ming Ding, Wenyi Hong, Jianqiao Wangni, Zhuoyi Yang, and Jie Tang.
\newblock Relay diffusion: Unifying diffusion process across resolutions for image synthesis.
\newblock \emph{arXiv preprint arXiv:2309.03350}, 2023.

\bibitem[Tschannen et~al.(2024)Tschannen, Pinto, and Kolesnikov]{tschannen2024jetformer}
Michael Tschannen, Andr{\'e}~Susano Pinto, and Alexander Kolesnikov.
\newblock Jetformer: An autoregressive generative model of raw images and text.
\newblock \emph{arXiv preprint arXiv:2411.19722}, 2024.

\bibitem[Vaswani et~al.(2017)Vaswani, Shazeer, Parmar, Uszkoreit, Jones, Gomez, Kaiser, and Polosukhin]{vaswani2017attention}
Ashish Vaswani, Noam~M. Shazeer, Niki Parmar, Jakob Uszkoreit, Llion Jones, Aidan~N. Gomez, Lukasz Kaiser, and Illia Polosukhin.
\newblock Attention is all you need.
\newblock In \emph{Neural Information Processing Systems}, 2017.

\bibitem[Wang et~al.(2024{\natexlab{a}})Wang, Dufour, Andreou, Cani, Abrevaya, Picard, and Kalogeiton]{cfg_analysis}
Xi Wang, Nicolas Dufour, Nefeli Andreou, Marie-Paule Cani, Victoria~Fern{\'a}ndez Abrevaya, David Picard, and Vicky Kalogeiton.
\newblock Analysis of classifier-free guidance weight schedulers.
\newblock \emph{arXiv preprint arXiv:2404.13040}, 2024{\natexlab{a}}.

\bibitem[Wang et~al.(2024{\natexlab{b}})Wang, Zhang, Luo, Sun, Cui, Wang, Zhang, Wang, Li, Yu, et~al.]{wang2024emu3}
Xinlong Wang, Xiaosong Zhang, Zhengxiong Luo, Quan Sun, Yufeng Cui, Jinsheng Wang, Fan Zhang, Yueze Wang, Zhen Li, Qiying Yu, et~al.
\newblock Emu3: Next-token prediction is all you need.
\newblock \emph{arXiv preprint arXiv:2409.18869}, 2024{\natexlab{b}}.

\bibitem[Yan et~al.(2024)Yan, Liu, Pan, Liew, qiang liu, and Feng]{yan2024perflow}
Hanshu Yan, Xingchao Liu, Jiachun Pan, Jun~Hao Liew, qiang liu, and Jiashi Feng.
\newblock Pe{RF}low: Piecewise rectified flow as universal plug-and-play accelerator.
\newblock In \emph{The Thirty-eighth Annual Conference on Neural Information Processing Systems}, 2024.

\bibitem[Yang et~al.(2024)Yang, Teng, Zheng, Ding, Huang, Xu, Yang, Hong, Zhang, Feng, et~al.]{yang2024cogvideox}
Zhuoyi Yang, Jiayan Teng, Wendi Zheng, Ming Ding, Shiyu Huang, Jiazheng Xu, Yuanming Yang, Wenyi Hong, Xiaohan Zhang, Guanyu Feng, et~al.
\newblock Cogvideox: Text-to-video diffusion models with an expert transformer.
\newblock \emph{arXiv preprint arXiv:2408.06072}, 2024.

\bibitem[Zeng et~al.(2022)Zeng, Vahdat, Williams, Gojcic, Litany, Fidler, and Kreis]{zeng2022lion}
Xiaohui Zeng, Arash Vahdat, Francis Williams, Zan Gojcic, Or Litany, Sanja Fidler, and Karsten Kreis.
\newblock {LION}: Latent point diffusion models for 3d shape generation.
\newblock In \emph{Advances in Neural Information Processing Systems}, 2022.

\bibitem[Zhai et~al.(2024)Zhai, Zhang, Nakkiran, Berthelot, Gu, Zheng, Chen, Bautista, Jaitly, and Susskind]{zhai2024normalizing}
Shuangfei Zhai, Ruixiang Zhang, Preetum Nakkiran, David Berthelot, Jiatao Gu, Huangjie Zheng, Tianrong Chen, Miguel~Angel Bautista, Navdeep Jaitly, and Josh Susskind.
\newblock Normalizing flows are capable generative models.
\newblock \emph{arXiv preprint arXiv:2412.06329}, 2024.

\bibitem[Zhang et~al.(2025)Zhang, Li, Chen, Ge, Sun, Zhang, Jiang, Yuan, Peng, and Luo]{zhang2025flashvideo}
Shilong Zhang, Wenbo Li, Shoufa Chen, Chongjian Ge, Peize Sun, Yida Zhang, Yi Jiang, Zehuan Yuan, Binyue Peng, and Ping Luo.
\newblock Flashvideo: Flowing fidelity to detail for efficient high-resolution video generation.
\newblock \emph{arXiv preprint arXiv:2502.05179}, 2025.

\bibitem[Zhou et~al.(2024)Zhou, Yu, Babu, Tirumala, Yasunaga, Shamis, Kahn, Ma, Zettlemoyer, and Levy]{zhou2024transfusion}
Chunting Zhou, Lili Yu, Arun Babu, Kushal Tirumala, Michihiro Yasunaga, Leonid Shamis, Jacob Kahn, Xuezhe Ma, Luke Zettlemoyer, and Omer Levy.
\newblock Transfusion: Predict the next token and diffuse images with one multi-modal model.
\newblock \emph{arXiv preprint arXiv:2408.11039}, 2024.

\end{thebibliography}
}

\end{document}